# Unsupervised Waste Classification By Dual-Encoder Contrastive Learning and Multi-Clustering Voting (DECMCV)

Kui Huang[1,2,#], Mengke Song[1,2,#], Shuo Ba[1,2], Ling An[1,2], Huajie Liang[1,2], Huanxi Deng[1], Yang Liu[1,3], Zhenyu Zhang[1,2,*], and Chichun Zhou[1,2,*]

[1]School of engineering, Dali university; Yunnan, 671003, China

[2]Air-Space-Ground Integrated Intelligence and Big Data Application Engineering Research Center of Yunnan Provincial Department of Education, Yunnan, 671003, China

[3]School of Electronic Information Engineering, Liuzhou Vocational &Technical College, Guangxi, 545006, China

[*]Corresponding author: Zhenyu Zhang: zhangzhenyu@dali.edu.cn

Chichun Zhou: zhouchichun@dali.edu.cn

[#]These authors contribute equivalently

Emails of other authors: Kui Huang: huangkui@stu.dali.edu.cn

Mengke Song: songmengke@stu.dali.edu.cn

Shuo Ba: bashuo@stu.dali.edu.cn

Ling An: anling@stu.dali.edu.cn

Huajie Liang: lianghuajie@stu.dali.edu.cn

Huanxi Deng: denghuanxi@stu.dali.edu.cn

Yang Liu: liuyang@stu.dali.edu.cn

## Key words

# Abstract

Waste classification is crucial for improving processing efficiency and reducing environmental pollution. Supervised deep learning methods are commonly used for automated waste classification, but they rely heavily on large labeled datasets, which are costly and inefficient to obtain. Real-world waste data often exhibit category and style biases, such as variations in camera angles, lighting conditions, and types of waste, which can impact the model's performance and generalization ability. Therefore, constructing a bias-free dataset is essential. Manual labeling is not only costly but also inefficient. While self-supervised learning helps address data scarcity, it still depends on some labeled data and generally results in lower accuracy compared to supervised methods. Unsupervised methods show potential in certain cases but typically do not perform as well as supervised models, highlighting the need for an efficient and cost-effective unsupervised approach. This study presents a novel unsupervised method, Dual-Encoder Contrastive Learning with Multi-Clustering Voting (DECMCV). The approach involves using a pre-trained ConvNeXt model for image encoding, leveraging VisionTransformer to generate positive samples, and applying a multi-clustering voting mechanism to address data labeling and domain shift issues. Experimental results demonstrate that DECMCV achieves classification accuracies of 93.78% and 98.29% on the TrashNet and Huawei Cloud datasets, respectively, outperforming or matching supervised models. On a real-world dataset of 4,169 waste images, only 50 labeled samples were needed to accurately label thousands, improving classification accuracy by 29.85% compared to supervised models. This method effectively addresses style differences, enhances model generalization, and contributes to the advancement of automated waste classification.

# 1. Introduction

Waste classification is critical for improving waste processing efficiency and reducing environmental pollution. By sorting waste, not only can recycling and processing efficiency be improved and costs reduced, but it can also promote resource recycling, such as converting kitchen waste into organic fertilizer. However, hazardous waste like old batteries, fluorescent light bulbs, and expired medications, if mixed with regular waste, can contaminate soil, water, and air, posing serious environmental and health risks (Abubakar et al., 2022; Naroznova et al., 2016; Zhou et al., 2015). Currently, traditional waste classification heavily relies on manual methods. While these methods meet accuracy and effectiveness requirements, they are inefficient and costly (Kaza et al., 2018; Robinson, 2007). Therefore, the development of efficient, automated waste classification methods that do not depend on human labor is key to solving these problems. One area of focus in this research is automated classification based on waste images.

Although some public datasets have been developed to align with general standards (Yang, 2019; Yang & Li, 2020; Wu et al., 2022), they still face many issues in the context of waste classification research and application. Due to factors such as regional, seasonal, and equipment differences, there are biases in both categories and styles between the training and test data, leading to degraded model performance (Yang & Li, 2020; Wu et al., 2022). For instance, datasets like TACO (Trash Annotations in Context) and TrashNet exhibit such issues, where the feature distribution in public datasets differs significantly from real-world environments (Yang, 2019; Yang & Li, 2020). The style differences within these domain shifts are particularly prominent. On one hand, significant discrepancies in collection environments and equipment play a major role. Public dataset images are often captured in cleaner environments and frequently sourced from the internet (Yang, 2019; Yang & Li, 2020), which contrasts sharply with the actual waste environments, thus interfering with model feature extraction. Different data collection methods result in variations in shape, brightness, and angles in similar waste images, leading to differences in features. Moreover, multispectral images captured with special equipment differ significantly from standard RGB images, making it difficult for models trained on RGB images to recognize multispectral features (Carrera et al., 2022; Shiddiq et al., 2023; Wu et al., 2022). On the other hand, public datasets fail to account for the temporal and spatial variations in waste types. In the same region, waste composition changes with the seasons, and across regions, it can vary significantly due to seasonal characteristics, residential habits, or the functional aspects of different areas (Chantou et al., 2013; Kumar et al., 2009). As shown in Figure 1, public datasets have limited coverage in terms of both scope and time, making it difficult to capture these variations. For example, the TrashNet dataset only includes six categories (Yang, 2019), which results in models struggling to recognize new waste categories, severely affecting classification performance.

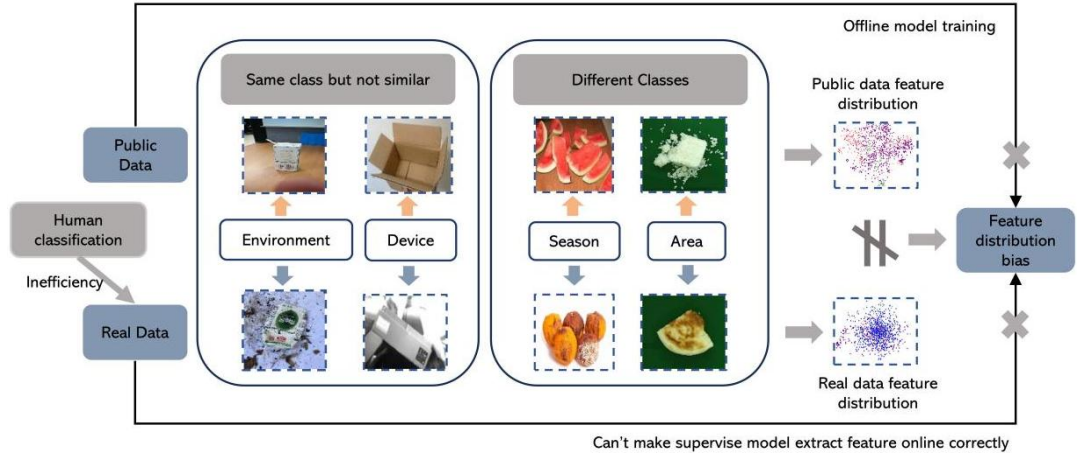

Figure 1. The contributing factors to the differences in feature distribution between public garbage datasets and real garbage data that lead to the insufficient generalizability of supervised models, as well as the constraints on the construction of real-world garbage datasets.

In the research and application of automated waste classification, supervised deep learning classification algorithms, such as CNN models, play a key role (Yu & Grammenos, 2021). These models offer advantages in classification speed and accuracy, and their powerful feature extraction capabilities make them highly effective for classification tasks. They are commonly used to optimize models and improve performance. However, supervised learning has significant drawbacks. On one hand, it heavily depends on manually labeled data, and creating task-specific labeled training datasets is costly. On the other hand, its performance deteriorates when faced with test datasets that exhibit domain differences. Even minor discrepancies between new data and the training data can significantly reduce classification accuracy. Therefore, constructing a universal training dataset that closely resembles real-world waste classification data is essential for achieving efficient automated classification through supervised learning.

In light of the challenges associated with supervised learning, self-supervised learning and unsupervised domain adaptation (UDA, see Appendix B) have emerged as key solutions to address these issues. Self-supervised learning focuses on learning general feature representations from unlabeled data by decoupling feature representation from label mapping, enabling effective data classification. Common techniques in self-supervised learning include autoencoders, contrastive learning, and feature extraction. In the context of waste classification, an EfficientNetv2 model trained with self-supervised learning is used for feature extraction, combined with Mask R-CNN for detection, yielding promising results. However, self-supervised learning is not without its limitations—it still partially depends on manually labeled data, which poses a significant barrier to its widespread application in large-scale waste classification scenarios. On the other hand, UDA aims to transfer knowledge from a labeled training set to unlabeled data with domain shifts, enabling accurate data classification. Techniques like Maximum Mean Discrepancy (MMD), Generative Adversarial Networks (GAN), and Knowledge Distillation (KD) are commonly used in UDA, with black-box unsupervised domain adaptation (BBDA) being particularly prevalent. Despite its advantages, UDA faces considerable challenges when applied to waste classification, especially due to the vast variety of waste types in real-world scenarios, making cross-category waste classification particularly difficult.

Addressing the urgent need for an unsupervised approach to construct unbiased datasets for waste classification, this study presents the DECMCV—an innovative method grounded in contrastive learning. DECMCV offers two notable contributions: Enhanced Encoding Efficiency: The framework employs a pretrained ConvNeXt model (Woo et al., 2023) to encode waste images, while the feature representations generated by Vision Transformer (ViT, Dosovitskiy et al., 2020) are incorporated as positive samples, improving the overall encoding performance. Multi-Cluster Voting Mechanism: To refine the clustering process, features are analyzed using three distinct clustering algorithms: K-means (Hartigan & Wong, 1979), AGG (Agglomerative Clustering, Kane, 1995), and BIRCH (Balanced Iterative Reducing and Clustering using Hierarchies, Zhang et al., 1996). Clusters with inconsistent outcomes are excluded, and final labels are assigned through a majority vote. This methodology addresses challenges associated with domain shift in waste classification, minimizing reliance on annotated data while improving labeling efficiency and model adaptability. By reducing manual intervention, DECMCV provides a scalable and practical solution for constructing robust, high-quality datasets.

Experimental results demonstrate that the proposed method achieves classification accuracies of 93.78% and 98.29% on the TrashNet and Huawei Cloud datasets, respectively, outperforming or rivaling traditional supervised approaches. Furthermore, for a real-world dataset containing 4,169 waste images, only 50 annotated samples were required to accurately label the entire dataset, significantly reducing annotation costs and improving the efficiency of constructing unbiased training datasets. Compared to supervised models trained on public datasets, this method improves classification accuracy in test environments by 29.85%, offering substantial support for practical applications of automated waste classification.

The main contributions of this study can be summarized as follows:

1) A novel unsupervised waste classification method has been proposed, eliminating the need for manual annotation. This method enables precise and efficient labeling of real-world waste datasets, leading to the construction of an unbiased training dataset. Models trained on this dataset demonstrate robust generalization ability in real-world scenarios, facilitating automated waste classification.

2) A comprehensive real-world waste dataset comprising 4,169 images, reflecting diverse environmental features, was constructed to validate the approach. Experimental results highlight the impact of domain shift on supervised models, providing valuable insights into domain adaptability. This advancement opens new possibilities for applying automated waste classification in practical settings, significantly enhancing its feasibility and relevance.

## 2. Datasets and Experimental Configuration

To demonstrate the impact of domain shift on supervised machine learning (SML) in waste classification, we selected two representative public datasets: the TrashNet dataset and the Huawei Cloud dataset. These datasets were used to construct training and testing set combinations. Additionally, to examine the domain shift between real-world waste datasets and public datasets, a realistic waste classification task environment was simulated. Real-world waste data were collected and used as the testing set, while public datasets served as the training set. This setup aimed to validate both the presence of domain shift and the effectiveness of the proposed method. Details of the experimental setup and methodology are provided in Chapter 4.

### 2.1 Selection of public data sets

To validate the existence of domain shift in public datasets and assess the effectiveness of the proposed method, two representative datasets were utilized: the TrashNet dataset (Yang, 2019) and the Huawei Cloud Waste Classification Challenge dataset (Yang & Li, 2020). TrashNet, one of the most widely used public datasets, contains 2,572 images across six categories. The Huawei Cloud dataset includes 14,802 images spanning four major categories and 43 subcategories. In Chapter 4, experiments were conducted to evaluate the impact of domain shift on supervised learning models by testing various combinations of training and testing datasets, constructed from these public datasets, where the testing data simulated unseen conditions. These experiments also demonstrated the high robustness of the proposed method in addressing domain shift. To facilitate comparisons of supervised models' classification results, the categories in both datasets were redefined based on the

Huawei Cloud dataset's taxonomy: recyclable waste, residual waste, kitchen waste, and hazardous waste. This standardization ensured consistency and highlighted the cross-domain classification performance.

## 2.2 Construction of real data sets

To validate the domain shift between publicly available datasets and real-world data, we constructed a dataset under conditions closely resembling real-world waste classification tasks. Specifically, waste samples were randomly collected from a local waste station, and a custom data acquisition system was designed. This system included a conveyor belt, a control mechanism, and an industrial RGB camera.

During data collection, waste was manually dispersed and evenly placed on the conveyor belt. When the waste entered a light-shielded enclosure, the conveyor belt paused, allowing the industrial camera to capture high-resolution RGB images (1024 × 1024 pixels, in JPG format). Detailed information about the data acquisition setup and procedures is provided in Appendix A. This system was designed to replicate the operational environment and data conditions of real-world waste classification tasks as closely as possible.

Based on the classification standards of the Huawei Cloud waste dataset, the collected waste images were categorized into three classes: recyclable waste, residual waste, and kitchen waste. A total of 4,169 images were collected. This dataset serves as a critical resource for analyzing domain shifts and validating the effectiveness of the proposed method in real-world scenarios.

# 3. Unsupervised Approach for Building Unbiased Datasets

Building on prior studies (Zhou et al., 2022; Liu et al., 2023; Gao et al., 2023; Dai et al., 2023; Fang et al., 2023; Qiu et al.,2024; Fang et al.,2025; Liu et al., 2025), this research proposes an innovative unsupervised machine learning framework to construct an unbiased training dataset for waste classification. This dataset is then employed to train supervised models capable of automatically classifying diverse test dataset combinations. The proposed methodology comprises three core steps to achieve unsupervised waste image classification:

1) Feature Extraction via Pretrained Models: Assuming waste images belong to a general domain, this study leverages pre-trained supervised models trained on large-scale, general-domain datasets (Zhou et al., 2022). These models are used to extract critical features from individual waste images effectively.

2) Enhanced Feature Representation Through contrastive learning: Contrastive learning is employed to refine feature representations by comparing differences across samples. Dual-model outputs are used to generate positive samples, enabling the extraction of key features through a contrastive learning approach. An unsupervised voting mechanism is then applied to cluster and annotate these features, producing unbiased data labels.

3) Improved Clustering Accuracy via Multi-Model Voting: To address the limitations of single-model clustering methods, a multi-model voting mechanism, inspired by previous work, is implemented to enhance clustering accuracy. The clustered data is subsequently validated through efficient manual inspection, resulting in a high-quality, unbiased waste classification dataset.

The constructed unbiased training dataset is subsequently used to train supervised models, enabling the automated classification of new waste datasets in real-world settings. This approach effectively bridges domain differences while ensuring robust classification performance across diverse scenarios.

## 3.1 Automatic Construction of an Unbiased Database

**Feature Extraction via Pre-trained Deep Learning Models:** To extract features from waste images, we use ConvNeXt, a convolutional neural network (CNN) designed with Vision Transformer-inspired training techniques (Woo et al., 2023), and ViT, a model leveraging self-attention for enhanced feature learning. Both models are pre-trained on ImageNet, a benchmark dataset containing 14 million images across 1,000 classes, ensuring their capability to extract meaningful and domain-relevant features. Each image in the dataset $\{x_1, x_2, \ldots, x_n\}$ are unified into these models, with their outputs unified into 2048-dimensional vectors using a linear layer, forming the feature set $e\{e_1, e_2, \ldots, e_n\}$. This results in a feature set $e=ConvNeXt(x)/ViT(x)$, where each sample is represented as $\{e_{i1}, e_{i2}, \ldots, e_{i2048}\}$, ensuring consistency in feature representation across the dataset.

**Unsupervised Feature Learning with Dual-Encoder Contrastive Learning:** Contrastive learning is an unsupervised learning approach that focuses on learning discriminative features by comparing the similarities and differences between data samples without relying on label information. The primary mechanism of contrastive learning involves constructing positive and negative sample pairs. In this framework, positive samples (similar samples) are pulled closer together in the feature space, while negative samples (dissimilar samples) are pushed farther apart. This spatial adjustment helps extract essential features from the data. A significant challenge in traditional contrastive learning lies in efficiently constructing high-quality positive sample pairs. To overcome this, we propose a novel solution leveraging a dual-encoder system. Each encoder is based on a different pretrained model, such as ConvNeXt or Vision Transformer (ViT). These encoders generate feature representations for the same data sample, and the similarity between these representations is used to define positive pairs. This dual-encoder strategy enhances the credibility of the positive sample pairs, filtering out redundant features and extracting more fundamental ones. It addresses the limitations of traditional contrastive learning methods in constructing positive pairs, leading to more efficient and accurate feature extraction. This approach forms a solid foundation for subsequent tasks like data classification and feature analysis, making it a powerful method in unsupervised learning pipelines.

**Unsupervised Clustering with Multi-Model Voting:** Unsupervised clustering have become a key area of focus in machine learning, allowing data to be categorized without the need for labels. These methods generally begin by randomly initializing cluster centers within the feature space and iteratively updating them to group data based on the similarity between each sample and its respective center. To address the challenge that single unsupervised models often fall short compared to supervised methods, this study introduces a multi-model voting mechanism for clustering. Specifically, the image features processed through contrastive learning are input into three commonly used clustering models: K-means (Hartigan & Wong, 1979), AGG (Kane, 1995), and Birch (Zhang et al., 1996). Each model independently assigns the data to 50 clusters. Since clustering labels from different models may vary, K-means labels are used as the standard, with the labels from the other two models being aligned to match this standard (ranging from 0 to 50). The voting mechanism is then applied: if all three models assign the same label to a data point, it is grouped with others of the same label. In cases of label disagreement, the data point is discarded. Lastly, the clustered data undergoes manual labeling, where each cluster is labeled according to the dominant label within that group, thus streamlining the process. This method significantly increases labeling efficiency. Given that most data points within a cluster belong to a single category, the labeling process becomes highly efficient. When more than 80% of the points in a cluster belong to one category, labeling speed is nearly the same as labeling a single image, offering a significant improvement over traditional manual methods, as shown in Figure 2.

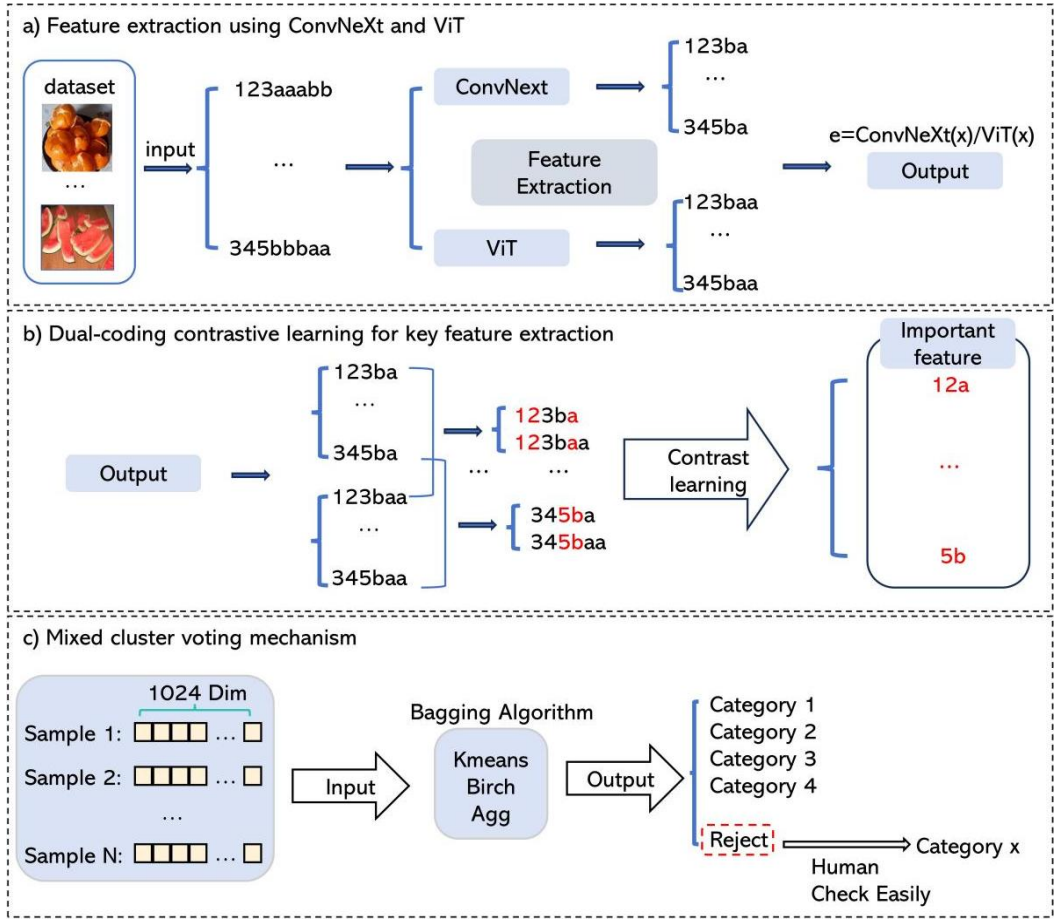

**Figure 2.** Flowchart of the Unsupervised Clustering Voting Algorithm: a) Feature extraction using ConvNeXt and ViT; b) Dual-coding contrastive learning for key feature extraction; c) Mixed clustering voting mechanism for enhanced accuracy.

### 3.2 Supervised model online classification based on unbiased datasets

Once trained, SML can store their parameters, enabling identical architectures to reuse them for inference tasks. This makes real-time online classification feasible using pre-trained models. To assess the online classification performance, the unbiased training dataset constructed in this study was utilized to train the widely recognized CNN model, GoogLeNet. Meanwhile, the discarded data served as the online test set. GoogLeNet, an advanced CNN architecture, is designed with a series of nine Inception modules. Each Inception module uses parallel convolutional paths with kernel sizes of $1 \times 1$, $3 \times 3$, and $5 \times 5$ to extract multi-scale features. These features are then integrated to expand the receptive field, allowing the network to capture a wider range of feature information (Szegedy et al., 2015).

## 4. Results and analysis

This section presents unsupervised labeling experiments conducted using both public datasets and the proposed dataset to validate the necessity and effectiveness of the three core steps in our methodology. To evaluate performance under varied conditions, different training-testing combinations were employed, including Huawei Cloud - TrashNet and Huawei Cloud + TrashNet - our dataset. The analysis of experimental results demonstrates the method's robustness in terms of accuracy, efficiency, and its ability to mitigate domain shifts. These findings confirm the validity of the approach and set the stage for the next sections, which will provide detailed discussions on experimental configurations and comprehensive analyses of dataset-specific results.

## 4.1 Experimental Parameter Configuration

For the dataset in Section 3.1, the data processed by unsupervised classification algorithms is designated as the training set, while the discarded data serves as the test set, with both being kept independent of each other. The training set is used for training supervised models, whereas the test set is employed for model evaluation after training. The experiments were conducted on a GTX-2070 server. In the feature encoding phase, the ConvNeXt model with pre-trained weights (ConvNeXt_xlarge_in22k) was used with an input size of 224*224, and the ViT model with pre-trained weights (B_16_imagenet1k) was used with an input size of 384*384. During the contrastive learning phase, the model was trained with a full dataset batch size, a learning rate of 0.001, and for 200 epochs. In the supervised training phase, the batch size was set to 16, with a learning rate of 0.001, and training was conducted for 200 epochs.

## 4.2 Results of Waste Classification on Different Datasets

This study evaluates the performance of both unsupervised clustering models and supervised classification models using three key metrics: Accuracy (Acc), Precision, and Recall. These metrics provide insights into overall classification effectiveness and the ability to correctly identify specific categories. The definitions are as follows: *Precision* $=N_1/N_2$. In this context, $N_1$ refers to the number of correctly classified images, while $N_2$ denotes the total number of images considered for classification. A higher accuracy indicates a greater proportion of correctly classified images in the overall results.

Recall is used to evaluate the classification performance for a specific category of images, and is defined as follows: *Recall* $=N_3/N_4$. Here, $N_3$ represents the number of correctly classified images for a particular category, and $N_4$ denotes the total number of images classified as that category. The higher the recall, the more images from that category are correctly identified.

**Classification Results on the Huawei Cloud Waste Dataset:** As shown in

Table 1, the unsupervised classification algorithm proposed in this study successfully classified 63.90% of the data in the Huawei Cloud waste dataset, achieving a classification accuracy of 98.29%. Compared to other supervised classification methods, our approach achieved the highest accuracy. Additionally, as shown in Figure 3-A, we reached an accuracy of 86.40% for hazardous waste classification, while the accuracy and recall rates for all other categories exceeded 92%.

**Table 1.** Comparison of results between our unsupervised algorithm and supervised method on Huawei Cloud data set

| Datasets | Training set | Test set | Test set accuracy | Data discarded |
|---|---|---|---|---|
| Our method | 0 | 11206 | 98.29% | 3584 (27.10%) |
| Cui et al.,2023 | 20000 | 4000 | 93.50% | 0 |
| Lin et al.,2021 | 20000 | 4000 | 95.62% | 0 |
| Fu et al.,2021 | 20000 | 4000 | 92.62% | 0 |

Figure 4-a illustrates the visualization of encoded data using the t-SNE algorithm (Van der Maaten & Hinton, 2008). Compared to the distribution of features derived from the raw image encoding, the encoded features obtained from large pretrained models exhibit distinct clustering for data points of the same class. This result highlights the superior feature extraction capability of large pretrained models. Moreover, after applying contrastive learning, the inter-cluster distances become significantly larger, further emphasizing the effectiveness of contrastive learning in extracting critical data features.

In the analysis of clustering outcomes, only two clusters out of 50 had a single-category proportion below 80%. This finding demonstrates that most clusters could be annotated nearly as quickly as labeling a single image, highlighting the high efficiency of post-hoc manual calibration in this study.

The algorithm presented in this study outperforms existing supervised methods in classification performance, as shown in Table 1, without relying on labeled data. Overall, our unsupervised classification algorithm efficiently extracts image category information and achieves high classification accuracy on the Huawei Cloud dataset, reaching a classification accuracy of 98.29% without the need for manual data labeling.

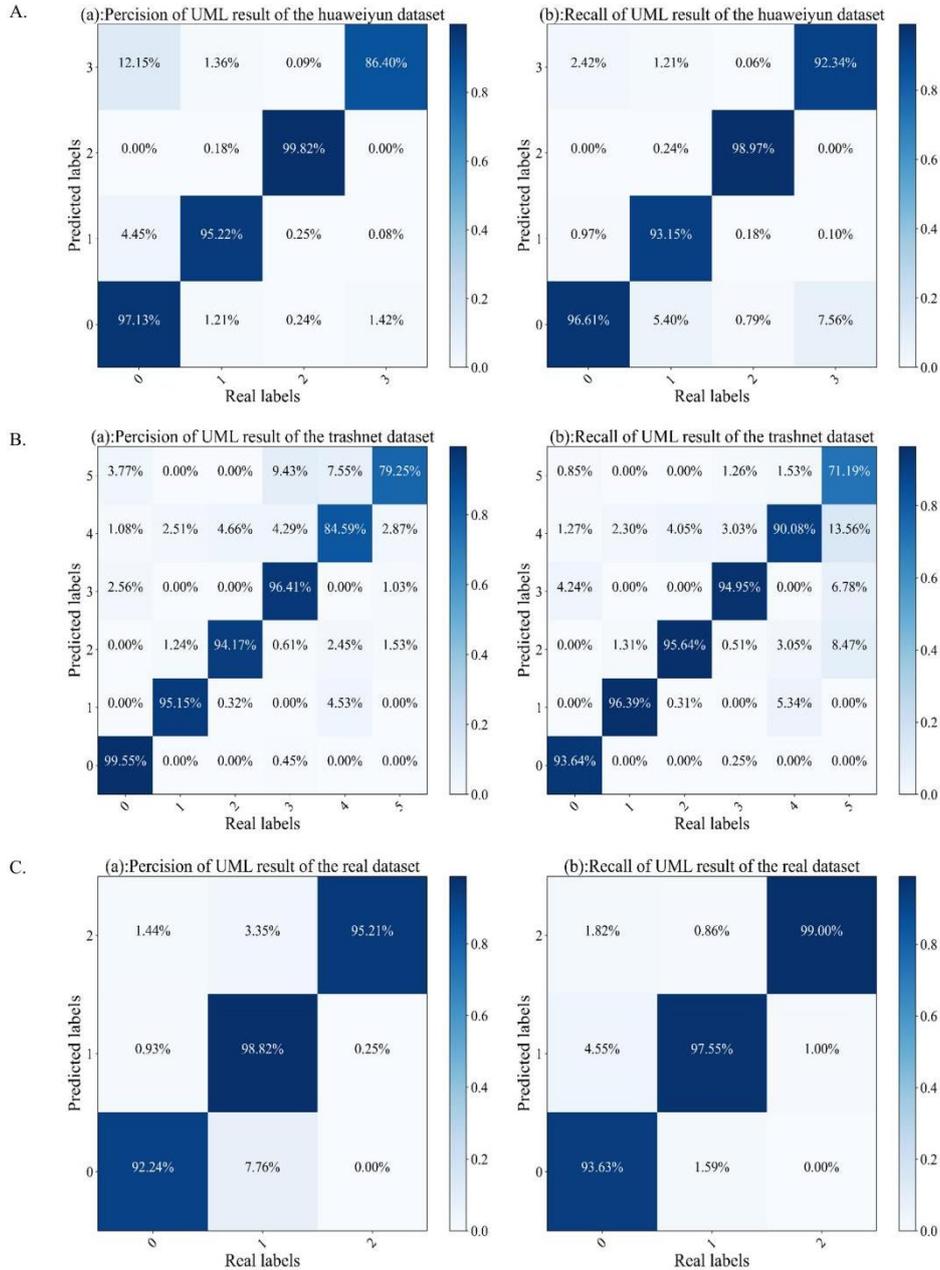

**Figure 3.** Illustrates the performance of the Unsupervised Classification Algorithm on three datasets. Panel A shows the results for the Huawei Cloud waste dataset. Subpanel A-(a) presents the Precison for each class, while Subpanel A-(b) displays the Recall for each class. The categories are as follows: category 0 represents recyclable waste, category 1 represents other types of waste, category 2 represents food waste, and category 3 represents harmful waste. Panel B presents the results for the Trashnet dataset. Subpanel B-(a) shows the Precision for each class, while Subpanel B-(b) illustrates the Recall for each class. The categories are defined as follows: 0 for cardboard, 1 for glass, 2 for metal, 3 for paper, 4 for plastic, and 5 for waste. Panel C displays the results for the dataset used in this study. Subpanel C-(a) shows the Precision for each class, while Subpanel C-(b) illustrates the Recall for each class. The categories are as follows: category 0 represents recyclable waste, category 1 represents other types of waste, and category 2 represents food waste.

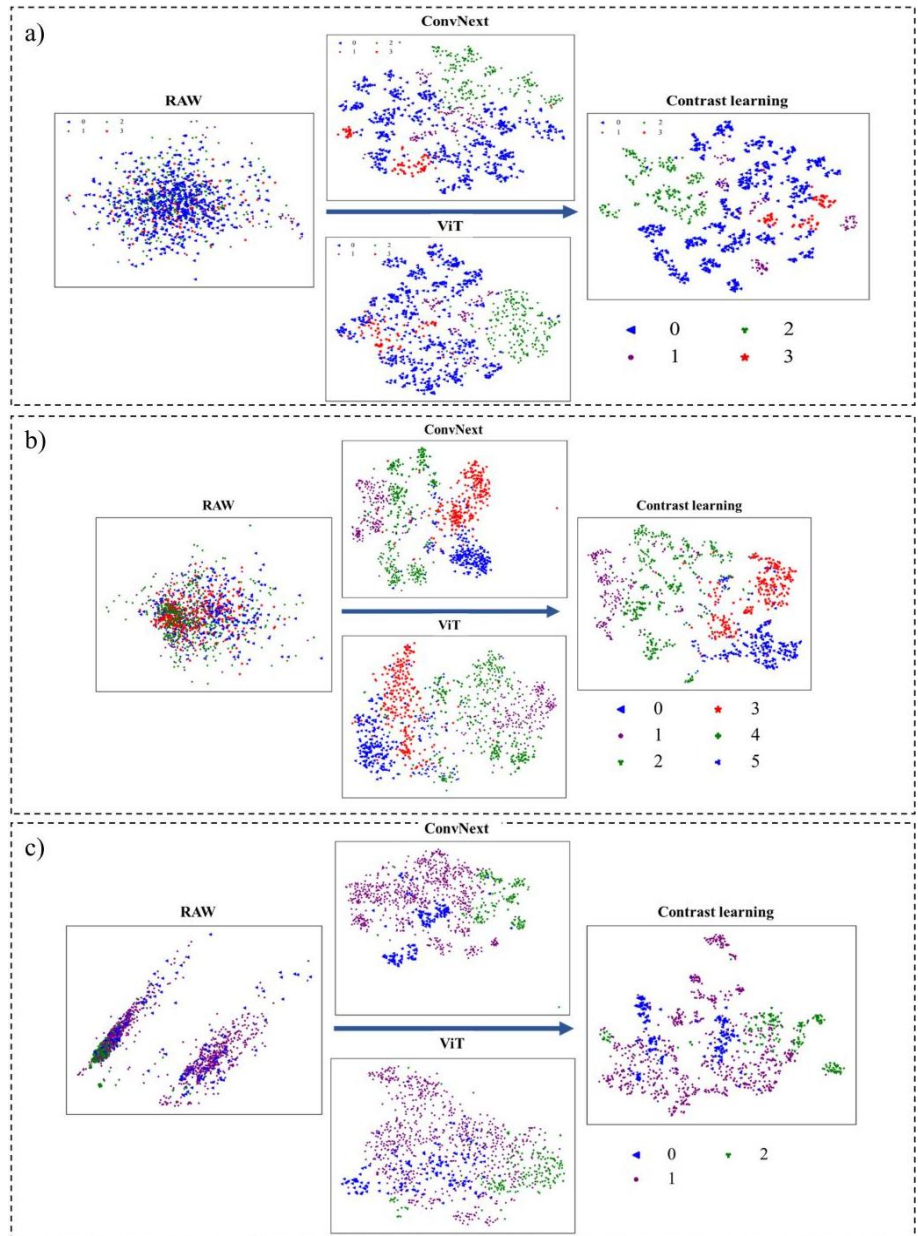

**Figure 4.** Visualization of the encoded information after each step of the unsupervised algorithm. In the legend a, category 0 represents the recyclable category, category 1 represents other categories, category 2 represents the food waste category, and category 3 represents the harmful category. In the legend b, 0 represents cardboard, 1 represents glass, 2 represents metal, 3 represents paper, 4 represents plastic, and 5 represents waste. In the legend c, category 0 represents recyclable categories, category 1 represents other categories, category 2 represents food waste categories.

**Evaluation of Classification Performance on the Trashnet Dataset:** As presented in Table 2, the Unsupervised Classification Algorithm proposed in this study successfully classified 63% of the data from the Trashnet dataset, achieving an overall accuracy of 93.78%.

**Table 2.** Evaluation and Comparison of Unsupervised Algorithm Performance vs. Supervised Methods on the Trashnet Dataset

| Datasets | Training set | Test set | Test set accuracy | Data discarded |
|---|---|---|---|---|
| GoogLeNet | 2057 | 515 | 80.02% | 0 |
| Our method | None | 1608 | 93.78% | 952 (37.00%) |
| Yu et al.,2020 | 2315 | 257 | 95.40% | 0 |
| Abu-Qdais et al.,2023 | 2064 | 508 | 96.06% | 0 |
| Ozkaya et al.,2019 | 1286 | 1286 | 97.86% | 0 |

As illustrated in Figure 3-B, the proposed Unsupervised Classification Algorithm achieved a Precision and Recall exceeding 94% for images categorized as cardboard, glass, metal, and paper, demonstrating the model's ability to accurately classify the majority of samples within these classes. For plastic images, the algorithm attained a precision of 84.59% and a recall of 90.08%, indicating strong performance in recognizing this category. However, the classification performance for waste images was comparatively lower, with a precision of 79.28% and a recall of 71.19%.

The relatively low classification performance for waste data, compared to the other five categories, may be attributed to the limited volume of data in this category, which prevents the unsupervised clustering algorithm from extracting sufficient class-specific features. Waste data accounts for only 5.42% of the total dataset, whereas the distributions for the other categories are more balanced.

To evaluate the effectiveness of the encoding and Principal Component Analysis (PCA) dimensionality reduction algorithm, we visualized the encoded feature data using the t-SNE algorithm. As shown in Figure 4-b, the large model encoding results exhibit a clear clustering structure, especially when compared to the original image encoding results. Furthermore, in the t-SNE plot after contrastive learning processing, the relative distances between clusters containing similar features are noticeably larger, which reflects the effectiveness of contrastive learning in feature extraction.

In the analysis of 50 clustering outcomes, only 7 categories had a class distribution below 80%, indicating that 86% of the clustering results could be quickly classified.

As detailed in Table 2, the algorithm proposed in this study outperforms traditional SML without relying on labeled data and achieves performance levels comparable to advanced SML. Overall, the Unsupervised Classification Algorithm demonstrated efficiency in extracting image class information from the Trashnet dataset, achieving high classification accuracy for most of the data without the need for manual labeling. The overall classification accuracy reached 93.78%.

**Results of Waste Classification on the Dataset Used in This Paper:** As shown in Table 3, the proposed Unsupervised Classification Algorithm successfully categorized 56.97% of the data in the real-world waste dataset, achieving a classification accuracy of 97.25%. Additionally, as illustrated in Figure 3-C, other evaluation metrics for the three waste categories all exceeded 92%, further demonstrating the effectiveness and robustness of the method.

Table 3. Results of our unsupervised algorithm on this paper's data set.

| Datasets | Our method |
| --- | --- |
| Data set for this article | 97.25% |
| Whether labeling is required | No |
| Data discard rate | 37.50% |

As depicted in Figure 4-c, we visualized the encoded data using the t-SNE algorithm. The results reveal that the combination of large model encoding and contrastive learning demonstrates exceptional capability in extracting critical features from the real-world waste dataset used in this study.

Overall, the proposed classification method performs effectively on the collected dataset, showcasing high efficiency and accuracy in classification tasks without relying on labeled data.

## 4.3 Ablation Analysis of Offline Classification

**The effect of large model coding:** As shown in Table 4, we applied the unsupervised voting-based classification algorithm to both the TrashNet dataset and the Huawei Cloud dataset using data encoded by large models and raw image pixel data.

The results indicate that the proposed method, leveraging ConvNeXt encoding and ViT encoding, achieves higher accuracy and lower discard rates compared to the method using raw image data. This demonstrates the effectiveness of large model encoding in extracting meaningful feature representations from images.

**The effect of contrast learning:** To achieve the goal of contrastive learning, which aims to bring similar features closer for same-class samples while enhancing the feature differences between different-class samples, it is essential to ensure that the samples retain the intrinsic features of their respective classes. The construction of positive and negative sample pairs is a critical issue. In this study, we employ a dual-encoding model to construct positive and negative sample pairs, allowing both large models to effectively extract the key features of the same-class data.

Table 4. A Comparative Analysis of Unsupervised Classification Results: Large Model

Encoding vs. Original Image Encoding on Trashnet and Huawei Cloud Datasets, Focusing on accuracy and Data Discard Rates.

| Result (Accuracy) | CovNeXt | ViT | RAW |
|---|---|---|---|
| Accuracy of the test set (Trashnet dataset) | 93.48% | 92.56% | 59.33% |
| Accuracy of the test set (Huawei Cloud dataset) | 97.80% | 96.74% | 61.58% |
| Result (The data drop rate) | CovNeXt | ViT | RAW |
| The data drop rate of Trashnet dataset | 37.50% | 47.00% | 55.00% |
| The data drop rate of Huawei Cloud dataset | 27.00% | 35.20% | 69.60% |

To validate the effectiveness of our method, we conduct a labeling test comparing the outputs of the large model's encoding and the encoding after processing through the contrastive model. As shown in Figure 5-a, the classification accuracy and drop rate of the encoded data after contrastive learning processing show a significant improvement over the results from the large model's encoding. This demonstrates that our positive and negative sample construction strategy allows the contrastive learning model to effectively learn the key features of the data.

**The effect of voting clustering algorithm:** To evaluate the effectiveness of the multi-model voting algorithm (Bagging), we used three clustering algorithms—K-means, Agg, and Birch—as participants in the voting process. As shown in Figure 5-b, the classification results from our multi-model voting algorithm achieved higher accuracy after discarding a subset of the data, indicating that our clustering voting mechanism can significantly enhance the accuracy of clustering results. Although some data were discarded after clustering voting, Figure 6 demonstrates that the supervised methods trained on the clustered data efficiently classify the discarded data. Furthermore, the clustering voting mechanism also improved post-labeling efficiency. For instance, in the best K-means clustering result for the Huawei Cloud dataset, the number of categories where less than 80% of images belong to a single class was twice as high as those in categories produced by the clustering voting algorithm.

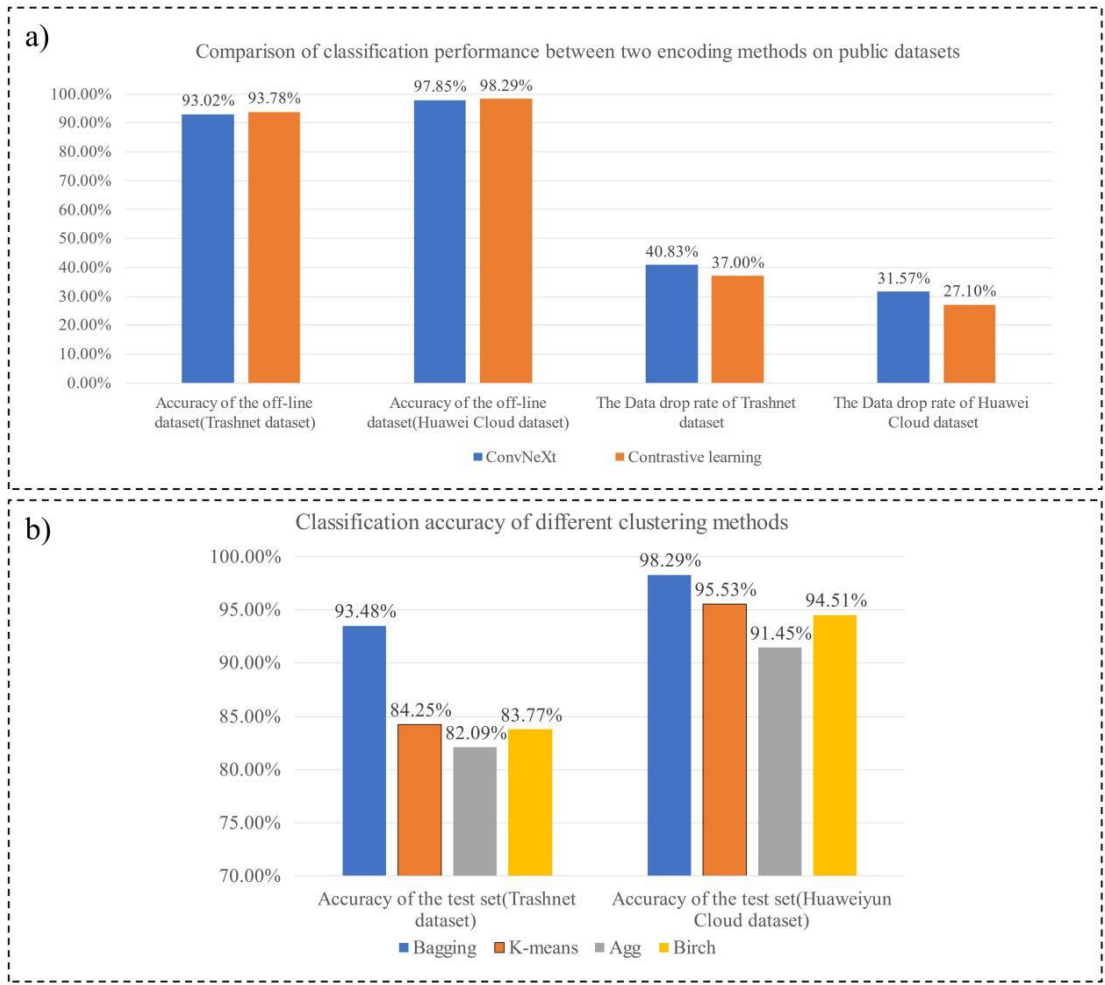

**Figure 5.** (a) shows the classification performance of the unsupervised algorithm using data processed by CL versus the original dimensional data; (b) presents the classification results of various CA applied to two public datasets

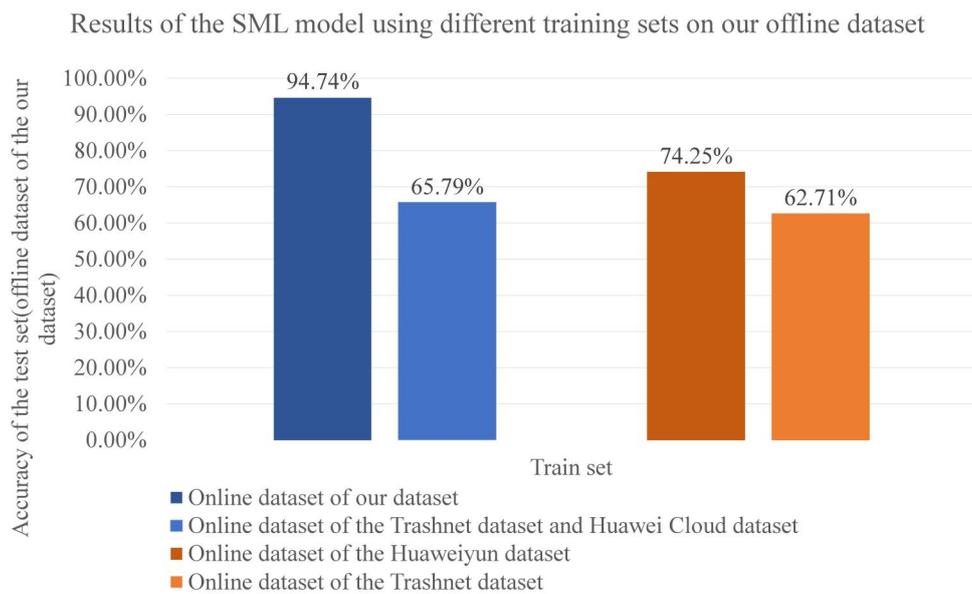

**Figure 6.** Comparison of Classification Accuracy Using Different Training Sets, presents a comparison of the classification accuracy results of supervised models trained on public

datasets and our own dataset, evaluated on the same test set. In this figure: Dark blue represents the classification performance of the supervised model when trained on the dataset from this study using unsupervised methods, with the test set consisting of discarded data. Light blue indicates the classification performance of the supervised model trained on a public dataset, evaluated on the same test set. Dark orange shows the classification performance of the supervised model trained on the Huawei Cloud dataset using unsupervised methods, tested on the Huawei Cloud data test set. Light orange represents the classification performance of the supervised model trained on the Trashnet dataset, evaluated on the same test set. This figure effectively highlights the impact of different training datasets on the classification performance of supervised models, demonstrating the robustness of the unsupervised pre-processing method used in this study.

## 4.4 Empirical Validation of the Unbiased Effectiveness

To demonstrate the impact of domain shift when using public datasets as general-purpose training datasets, and to showcase the robustness of our method against domain shift, we conducted two ablation experiments:

1) We trained a supervised model on the Trashnet dataset and the Huawei Cloud dataset separately, using the data processed by our method from each as the training sets. The discarded data from the Huawei Cloud dataset processed by our method, was used as the test set. We then compared the classification performance of the supervised models trained on each of these two datasets when evaluated on the same test set.

2) We created an open dataset consisting of both the Trashnet dataset and the Huawei Cloud dataset, and also used the data processed by our method as the training set. The discarded data from the proprietary dataset processed by our method was used as the test set. We compared the classification performance of supervised models trained on these two different training datasets, evaluated on the same test set.

**Performance Comparison of the Proposed Method and Traditional Supervised Models on Public Datasets:** This study utilizes the Trashnet Dataset and the Huawei Cloud Dataset, processed and labeled by our method, as training sets for the supervised model. The test set comprises discarded data from the Huawei Cloud Dataset, which is independent of the training data. As shown on the right side of Figure 6, the classification performance of the supervised model trained with the Huawei Cloud Dataset surpasses that of the supervised model trained with the Trashnet Dataset by 24.25%. This result highlights the presence of feature distribution discrepancies, also known as domain shifts, between different public datasets. Furthermore, the supervised model trained using data processed by our method achieves an accuracy of 81.71% on the test set.

In summary, these findings demonstrate that public datasets often fail to account for domain shifts, and that supervised models trained with data processed by our method can effectively classify discarded data. This approach enables high-accuracy annotation of the entire dataset.

**Comparison of the Proposed Method and Traditional Supervised Algorithms on this article's data set:** Additionally, we compared the performance of supervised models trained using two different training sets: a combined public dataset (Trashnet Dataset and Huawei Cloud Dataset) and a real-world waste dataset processed by our method. The test set consisted of discarded data from the real-world waste dataset processed by our method. As shown on the left side of Figure 6, the supervised model trained with our dataset outperformed the one trained with the public dataset by 28.95% in classification accuracy. This result highlights a clear feature distribution discrepancy, or domain shift, between public datasets and real-world waste data. Furthermore, the supervised model trained with our method's dataset achieved a classification accuracy of 94.74% on the test set.

In conclusion, the experiments confirm the presence of domain shift between public and real-world datasets. They also demonstrate that in practical waste classification tasks, supervised models trained with datasets processed by our method can effectively classify discarded data with high accuracy, mitigating the effects of domain shift and enabling precise dataset annotation.

## 5. Conclusion and discussion

This study proposes a fast labeling method for waste datasets based on unsupervised learning, aiming to address the domain shift issue caused by discrepancies in feature distributions between public waste datasets and real-world waste classification data, a common challenge in traditional supervised learning. By leveraging large pre-trained models (ConvNeXt and ViT) for feature extraction, and combining contrastive learning with a multi-model voting algorithm, the proposed method demonstrates excellent performance in labeling tasks on both public datasets like Huawei Cloud and TrashNet, as well as the dataset used in this study. Experimental results show that our method not only surpasses traditional supervised learning methods in terms of accuracy but also exhibits strong robustness across different datasets, successfully mitigating the impact of domain shift on supervised learning models.

To further investigate the impact of domain shift, this study first analyzes the feature distribution differences between public datasets and real-world waste data. Experimental validation shows that the supervised model trained with our approach demonstrates strong robustness in the presence of domain shift, and its effectiveness is proven in classifying real-world online datasets. These results indicate that our method effectively mitigates the decline in classification accuracy caused by domain shift, thereby enhancing the generalization ability of supervised learning models across different datasets.

We conducted ablation experiments on public datasets to evaluate the contribution of the three core components of our approach. The findings underscore the distinct role each technique plays in enhancing overall performance. Specifically, large model encoding efficiently captures essential image features; contrastive learning refines these features, enhancing their representational accuracy; and the clustering voting algorithm significantly improves the precision of clustering outcomes, ensuring robust and reliable classification results.

Overall, the method proposed in this study enables fast and accurate waste classification without the need for manual labeling of each individual item. Additionally, our approach demonstrates strong performance in handling domain shift, particularly in addressing the issue of feature distribution discrepancies between the training set and real-world data. It effectively reduces the negative impact of such discrepancies on the generalization ability of supervised models. Although the unsupervised algorithm still requires visual inspection and correction for 50 clustering categories, the labeling efficiency is nearly equivalent to labeling a single image at a time, as most data within each cluster belongs to the same category. Therefore, by labeling just 50 images, we can accurately label tens of thousands of samples, significantly improving labeling efficiency.

In real-world waste classification tasks, the method proposed in this study facilitates the rapid and accurate automatic labeling of waste data collected from diverse environmental settings. This capability ensures that, even as waste categories evolve over time or across regions, the method remains effective in annotating new data with high precision. By aligning the feature distributions between the training domain and the actual data domain, this approach effectively mitigates domain shift, thereby enhancing the generalization ability of the supervised learning model. As a result, the proposed method holds significant potential for supporting the widespread deployment of supervised learning model in various waste classification stations, offering substantial practical value.

# Appendix

## Appendix A: Real-World Waste Classification Dataset Construction

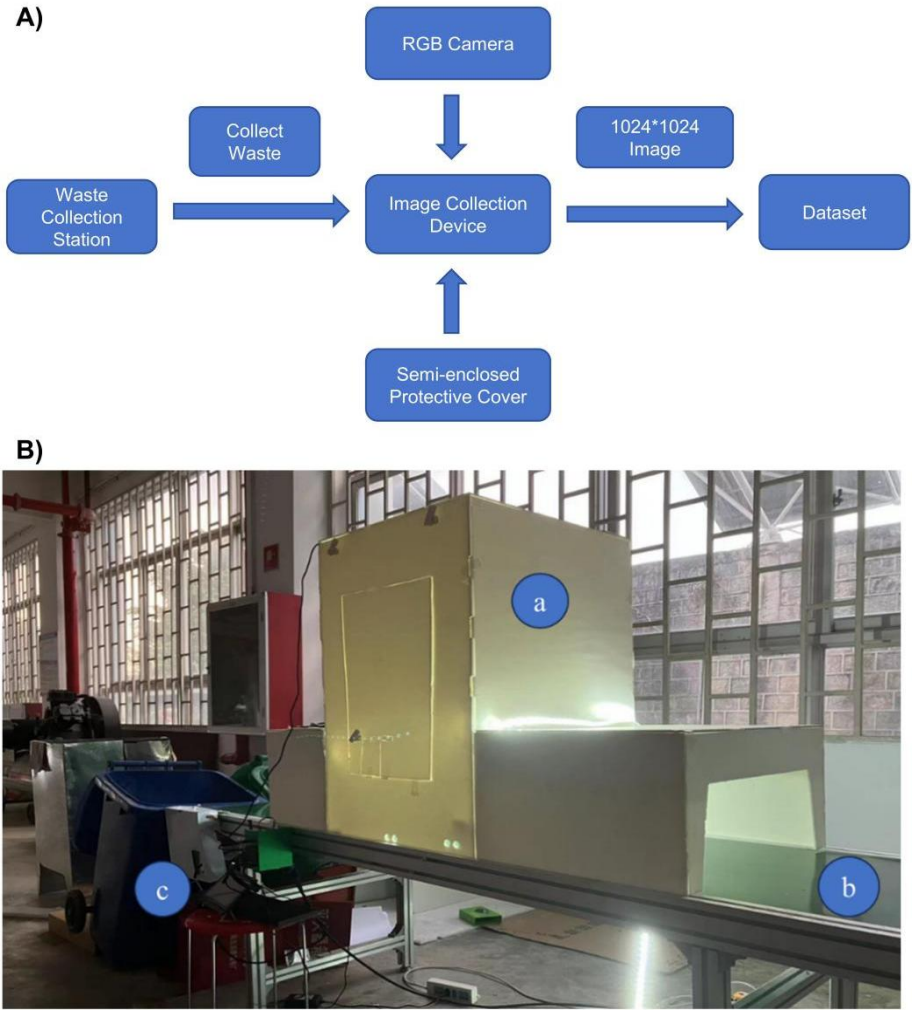

**Figure 7.** (A) illustrates the waste data collection process. (B) shows the components of the collection facility, where (B-a) represents the semi-enclosed protective cover, (B-b) the conveyor belt, and (B-c) the control system.

## Appendix B: Unsupervised Domain adaptation

Unsupervised Domain Adaptation (UDA) is a widely adopted transfer learning strategy that bridges the distributional gap between the source domain (SD) and target domain (TD). By leveraging feature extraction and adversarial training, UDA employs unlabeled data from the TD to reduce distributional discrepancies, enhancing the generalization capability of the model. In particular, Domain Adversarial Training facilitates the alignment of domain-invariant feature representations by addressing environmental inconsistencies such as lighting and color differences. However, growing categorical and stylistic differences between the SD and TD, coupled with variations in data acquisition conditions, exacerbate adaptation challenges. For instance, inconsistencies in waste categories and image-capturing environments can result in significant distributional gaps. In cases where labeled TD data is scarce, UDA enables knowledge transfer from the SD, narrowing the distributional gap and improving model performance on the TD.

To mitigate the impact of domain shift on model performance, researchers have explored UDA techniques, such as adversarial adaptation methods (e.g., Domain-Adversarial Neural Networks, DANN) and contrastive learning frameworks (e.g., Maximum Mean Discrepancy, MMD). While effective, these approaches face challenges in addressing class mismatches and style inconsistencies between the SD and TD. For example, the SD may consist of a waste classification dataset from a specific city, featuring common waste types captured under uniform conditions. In contrast, the TD could be from another region, where waste categories and environmental factors vary significantly, causing a distribution gap between domains. In such cases, the model needs to adapt to new categories and environmental conditions. One possible solution is domain-specific fine-tuning, which involves pretraining a model on public datasets and refining it using limited labeled data from the TD. While this approach helps reduce labeling efforts, it may struggle when encountering novel waste characteristics not covered during pretraining.

Domain adaptation faces a persistent challenge in achieving style alignment, requiring the model to ignore superficial visual differences and focus instead on the intrinsic characteristics of waste categories. To address this, methods such as online learning and incremental learning have been developed, allowing models to adapt dynamically to new data post-deployment. While these approaches reduce the need for re-labeling, they are highly sensitive to noisy data, which can destabilize model updates. For example, noise introduced by sensor errors or environmental disruptions during waste image collection may impair the model's ability to accurately extract waste-specific features, ultimately degrading classification accuracy and stability.

In summary, while domain adaptation techniques effectively bridge environmental differences between the SD and TD, significant challenges remain, particularly in addressing category misalignment, style inconsistencies, and the impact of noisy inputs.

## Acknowledgment

This work was supported by the National Natural Science Foundation of China (42367066, 62106033), Yunnan Fundamental Research Projects (202401AT070016, 202301BA070001-037), National Observation and Research Station of Erhai Lake Ecosystem in Yunnan (2022ZZ01), Yunnan Province Dali Prefecture Science and Technology Bureau Social Development Field Project (20232904E030002).

## Author contributions statement

Conceptualization, Chichun Zhou, Zhenyu Zhang; Methodology, Kui Huang, Shuo

Ba. Huajie Liang, Huanxi Deng, Yang Liu and Ling An; Formal analysis, Chichun Zhou, Zhenyu Zhang, Yang Liu, Ling An and Shuo Ba; Writing - original draft, Kui Huang, Mengke Song, Shuo Ba, Huanxi Deng and Chichun Zhou; Reviewing the manuscript, ChichunZhou, Zhenyu Zhang, Ling An, and Yang Liu; All the authors have read and agreed to the published version of the manuscript.

## Additional information

Conflicts of Interest: The authors declare that they have no known competing financial interests or personal relationships that could have appeared to influence the work reported in this paper.